\documentclass{article}

\usepackage{multicol}
\usepackage{graphicx, caption, subcaption}

    \usepackage[final, nonatbib]{neurips_2019_ml4ps}

\usepackage[utf8]{inputenc} 
\usepackage[T1]{fontenc}    
\usepackage{hyperref}       
\usepackage{url}            
\usepackage{booktabs}       
\usepackage{amsfonts}       
\usepackage{nicefrac}       
\usepackage{microtype}      

\title{Accounting for Physics Uncertainty in Ultrasonic Wave Propagation using Deep Learning}

\author{
  Ishan D. Khurjekar \And
   Joel B. Harley \And\\
  Department of Electrical and Computer Engineering,\\
  University of Florida, \\ 
  Gainesville, FL 32601 \\
  \texttt{ishan.khurjekar@ufl.edu, joel.harley@ufl.edu} \\
}

\begin{document}

\maketitle

\begin{abstract}
  Ultrasonic guided waves are commonly used to localize structural damage in infrastructures such as buildings, airplanes, bridges. Damage localization can be viewed as an inverse problem. Physical model based techniques are popular for guided wave based damage localization. The performance of these techniques depend on the degree of faithfulness with which the physical model describes wave propagation. External factors such as environmental variations and random noise are a source of uncertainty in wave propagation. The physical modeling of uncertainty in an inverse problem is still a challenging problem. In this work, we propose a deep learning based model for robust damage localization in presence of uncertainty. Wave data with uncertainty is simulated to reflect variations due to external factors and Gaussian noise is added to reflect random noise in the environment. After evaluating the localization error on test data  with uncertainty, we observe that the deep learning model trained with uncertainty can learn robust representations. The approach shows potential for dealing with uncertainty in physical science problems using deep learning models.
\end{abstract}

\section{Introduction}

With the exponential increase in the number of civil infrastructures, including buildings and bridges, it has become imperative to monitor their structural integrity. A number of techniques exist for monitoring structural health. Ultrasonic guided wave testing (UGWT) is one such popular, non-destructive method. Ultrasonic guided waves can scan large areas and are sensitive to damage. Hence they are a popular option for damage localization systems. An UGWT setup consists of a spatial array of sensors that can transmit and / or receive acoustic signals. Based on these acoustic signals, multiple techniques have been developed for damage localization \cite{michaels2008detection}.

Wave physics based techniques are a common choice for solving the inverse problem of guided wave based localization \cite{fan2011vibration}. Typically, such techniques for damage localization use a theoretical model of wave propagation. By comparing the signal received at receiver sensors with the output of the physical model at all of the possible damage locations, the structural damage is localized. Yet, there exists uncertainty in guided wave propagation because of variations of external factors, such as temperature, humidity, air pressure and random noise \cite{sohn2006effects}. This uncertainty is challenging to incorporate in a physical wave propagation model because of the complex and dispersive nature of guided waves \cite{friswell2006damage}. 

To tackle these uncertainties, researchers have explored data-driven approaches for damage localization with guided waves \cite{harley2017managing}. Simultaneously, there has been a growing interest in machine learning models for physical parameter estimation problems \cite{raissi2019physics}\cite{karpatne2017physics}. To physically model real world phenomena in a system, tackling sources of uncertainty due to external factors is critical \cite{oberkampf2004challenge}. Accurate physical characterization of uncertainties in real world scenario for further use in machine learning models remains a challenging and open research problem.

In this work, we investigate the potential for tackling uncertainty in guided wave propagation using deep learning models. We simulate guided wave data with uncertainty in velocity of wave. We also simulate random noise by adding Gaussian noise to the data. We then build a deep neural network (DNN) that learns representations which are robust to the uncertainties in the data. We further motivate the use of dropout regularization as a tool to tackle uncertainty in physics inspired machine learning models. We validate our results on test data-set and conclude our discussion with future scope.

\section{Model-based damage localization and challenges}
\label{gen_inst}
\subsection{Damage localization setup}
Lamb waves are a specific case of guided waves commonly used in guided wave based damage localization algorithms \cite{worden2001rayleigh}. The far-field Lamb wave model is given by
\begin{equation}\label{eq:1}
    X(\omega_{q}) = \sum_{n}\sqrt \frac{1}{\kappa_{n}(\omega_{q})r}e^{-j\kappa_{n}(\omega_{q})r} 
\end{equation}
where X$(\omega_{q})$ in (\ref{eq:1}) is the frequency domain representation of the signal and is modeled as the summation across $n$ wave modes. The function $\kappa_{n}(\omega_{q})$ is the frequency and mode dependant wave-number (known as the dispersion relation) and $r$ is the distance travelled by the wave.

Consider the following setup for damage localization. An array of sensors is placed on a grid of dimensions $L \times W$. Every unique sensor pair acts as a transmitter-receiver pair. There exists a damage at some point in the grid that needs to be localized. The physical model for wave propagation in (\ref{eq:1}) is calculated for all possible damage locations on the grid. The location at which maximum correlation is obtained between the physical model and received data is the location estimate.

\subsection{Challenges with model-based damage localization}
\begin{figure*}[b]
  \begin{subfigure}[b]{0.45\textwidth}
  \centering
  \captionsetup{justification=centering}
    \includegraphics[width=\textwidth]{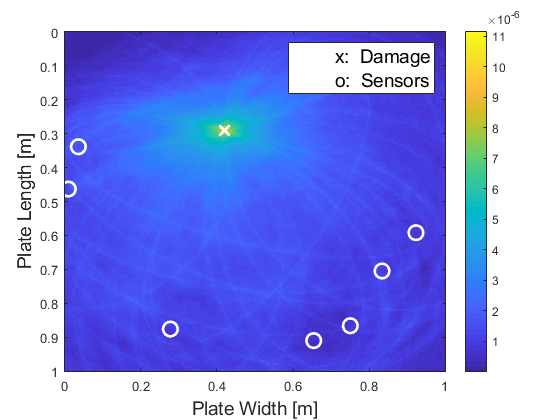}
    \caption{}
    \label{fig:1a}
  \end{subfigure}
  \hfill
  \begin{subfigure}[b]{0.45\textwidth}
  \centering
    \includegraphics[width=\textwidth]{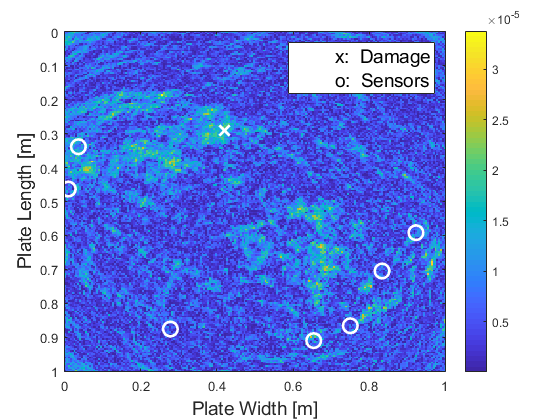}
    \caption{}
    \label{fig:1b}
  \end{subfigure}
  \caption{Localization heat-maps: (a) in noiseless conditions and (b) in noisy conditions (illustrated here with a signal-to-noise ratio (SNR) of 5 dB).}
\end{figure*}
 The dispersion relation between $\kappa$ and $\omega$ is critical to understanding the behaviour of waves in structures. This relation is dependent on the properties of the material in which the wave propagates. The accurate recovery of dispersion relation thus becomes complicated in the presence of external uncertainties such as temperature, humidity, air pressure and random noise which affect material properties. We observe that uncertainty in the dispersion relation leads to uncertainty in (\ref{eq:1}). Moreover, as group velocity (velocity of a wave packet) is given by 
\begin{equation}\label{eq:4}
    \nu_{g} = \frac{\partial \omega}{\partial \kappa}
\end{equation}
uncertainty in dispersion relation also leads to uncertainty in velocity. Similarly, random noise also affects the performance of model based localization. Figure 1 (a) and Figure 1 (b) show the performance in ideal (noiseless) conditions and noisy conditions, respectively, as a heat-map. The preceding discussion motivates our research direction of dealing with uncertainty in wave propagation.
\section{Deep neural network based localization framework}
\label{headings}
\subsection{Simulation setup}
\begin{figure}[htb]\label{f2}
    \includegraphics[width=14.0cm]{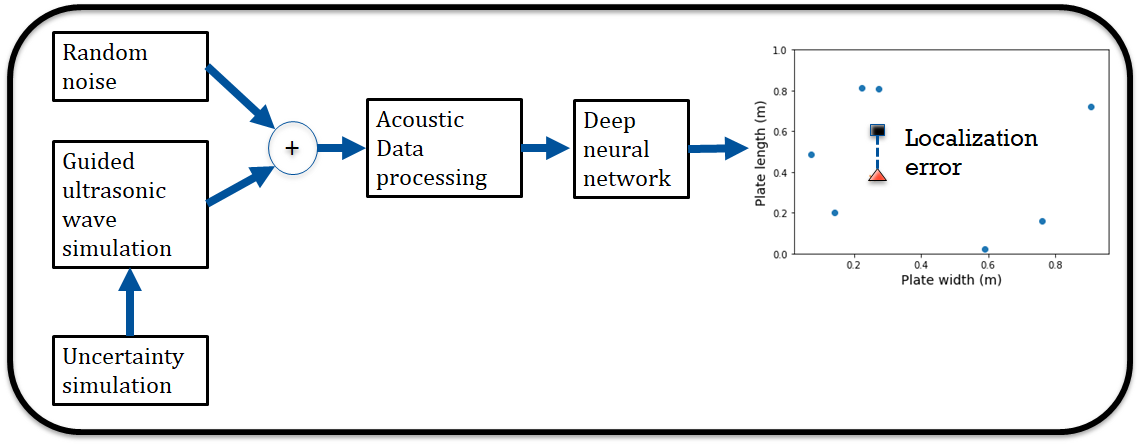}
    \caption{Simulation validation procedure }
    \label{fig:f2}
\end{figure}
For damage localization, we assume a plate with unit dimensions and the damage modeled at a random point. The damage acts as a point scatterer of incident waves. We use a sparse-array configuration for the sensors. We use random sensor placement to unbias the results based on sensor configuration. We simulate guided waves using (\ref{eq:1}) with $m = 8$ sensors ($M = 56$ unique sensor pairs) placed at random locations on the plate and $Q = 1000$ equally spaced frequencies from 0 to 1000 kHz. 

We refer to the $Q \times M$ time-domain wave data matrix and the corresponding damage location as one input-label pair. We do $t = 2500$ simulations to build a guided wave data-set with $2500$ samples. We divide this data-set into train and test set. As shown in Figure 2, we add uncertainty and random noise to the simulated wave data. We introduce a random multiplicative effect $\alpha$ on the dispersion curves such that modified dispersion curves are defined by $\kappa(\omega) = \alpha \kappa(\omega)$. This creates velocity uncertainty in the simulations. The random variable $\alpha$ is randomly sampled from a Gaussian distribution truncated between $0.7$ and $1.3$ and a standard deviation of $1$. The random noise is modeled as additive white Gaussian noise (AWGN). 

Next, the time-domain wave data is pre-processed appropriately for the DNN to ensure optimum performance. This includes standardizing it and flattening the matrix into a 1D vector to be used as input to the DNN. The DNN has 3 fully connected hidden layers. First hidden layer has $h_{1}$ = 300 nodes, second hidden layer has $h_{2}$ = 200 nodes, and third hidden layer has $h_{3}$ = 50 nodes. The output layer has 2 nodes, one each for $x$ and $y$ dimension localization. We choose loss function as the Euclidean distance between prediction of DNN and the actual damage location as shown in the extreme right of Figure 2. We train the DNN using Keras package \cite{chollet2015keras} for 50 epochs.

\subsection{How does DNN help tackle uncertainty in guided wave propagation ?}
The DNN has fully connected layers (every node from previous layer is connected to every node in the next layer). At the input layer, this enables the network to learn cross-frequency relationships for the dispersive wave data. Having multiple hidden layers in the DNN, likely also helps to learn more complex representations for the inverse mapping between the wave data and the damage location.

For a physical science problem, over-fitting is analogous to having an over-complex model that explains the available data but does not explain well on unseen data. We want to ensure that the network does not learn overly complex representations yet it should also account for the uncertainty. Over-fitting is tackled in machine learning community using regularization techniques such as dropout \cite{srivastava2014dropout}. When using dropout, nodes in the DNN are randomly dropped out of the network while training, which is equivalent to learning multiple models. Researchers have presented theory that casts dropout as a measure of model uncertainty \cite{gal2016dropout}. This motivates our research direction of using deep learning models and tools to deal with uncertainty in wave propagation.

\section{Results}
\label{Results}
The performance of the localization algorithms is quantified with average localization error (ALE) on the test data-set (wave data with uncertainties and random noise)
\begin{equation}\label{eq3}
    ALE = \frac{1}{T}\sum_{t = 1}^{T}\sqrt{(x - \overline{x})^{2} + (y - \overline{y})^{2}}
\end{equation}
where T is the number of samples in the data-set and $(x,y)$, $(\overline{x},\overline{y})$ are the actual and predicted damage locations respectively. 
\begin{figure*}[h]
  \begin{subfigure}[h]{0.45\textwidth}
  \centering
  \captionsetup{justification=centering}
    \includegraphics[width=\textwidth]{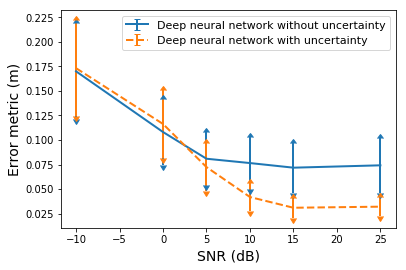}
    \caption{}
    \label{fig:3a}
  \end{subfigure}
  \hfill
  \begin{subfigure}[h]{0.45\textwidth}
  \centering
  \captionsetup{justification=centering}
    \includegraphics[width=\textwidth]{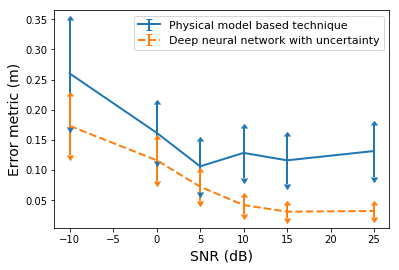}
    \caption{}
    \label{fig:3b}
  \end{subfigure}
  \caption{(a) Localization performance comparison of DNN trained with uncertainty $+$ 25 dB AWGN and DNN trained on ideal data (b) Localization performance comparison of DNN trained with uncertainty $+$ 25 dB AWGN and physical model based technique.}
  \end{figure*}

Figures 3 ((a)-(b)) are reported for test data-set (wave data with uncertainties and random noise). The x-axis represents the signal-to-noise ratio of the samples in the test data-set and y-axis represents the localization error metric (arrow-heads representing the standard deviation). Figure 3 (a) compares the localization performance of DNN trained with uncertainty and random noise with that of DNN trained on ideal data. At 25 dB, DNN's trained with and without uncertainty have localization errors of 0.0321 m and 0.0742 m respectively. Figure 3 (a) thus illustrates that model trained with uncertainty is able to learn representations that are robust to the uncertainty better than a model trained without any explicit uncertainty in the SNR range of 5 to 25 dB.

Figure 3 (b) compares the localization performance of the DNN trained with uncertainty and random noise with that of a physical model based technique \cite{friswell2006damage}. The physical model based technique uses a known wave propagation model which does not reflect the uncertainty present in a realistic setup. This leads to a highly variable performance trend for model based technique when tested on data with uncertainty. Figure 3 (b) illustrates that the DNN trained with uncertainty and noise has a superior performance compared to the physical model based technique.


\section{Conclusions}
\label{Conclusions}
We discussed the challenges posed by uncertainty due to external factors and random noise in ultrasonic wave based damage localization. We simulated wave data with uncertainty in velocity to reflect wave propagation in a realistic scenario. We further modeled random noise in environment as Gaussian noise. We trained a DNN on this simulated data and validated the simulation results on a test data-set. 

Based on the initial results, we can conclude that deep learning models can help deal with physical uncertainty in ultrasonic wave propagation. These results also provide further motivation for research of deep learning models and tools as a way of incorporating uncertainty in physical science problems.

\section{Acknowledgments}
This research is supported by the Air Force Office of Scientific Research under award number FA9550-17-1-0126 and the National Science Foundation under award number 1839704.


\bibliographystyle{IEEE}
\bibliography{references}

\begin{thebibliography}{10}

\bibitem{michaels2008detection}
Jennifer~E Michaels,
\newblock ``Detection, localization and characterization of damage in plates
  with an in situ array of spatially distributed ultrasonic sensors,''
\newblock {\em Smart Materials and Structures}, vol. 17, no. 3, pp. 035035,
  2008.

\bibitem{fan2011vibration}
Wei Fan and Pizhong Qiao,
\newblock ``Vibration-based damage identification methods: a review and
  comparative study,''
\newblock {\em Structural Health Monitoring}, vol. 10, no. 1, pp. 83--111,
  2011.

\bibitem{sohn2006effects}
Hoon Sohn,
\newblock ``Effects of environmental and operational variability on structural
  health monitoring,''
\newblock {\em Philosophical Transactions of the Royal Society A: Mathematical,
  Physical and Engineering Sciences}, vol. 365, no. 1851, pp. 539--560, 2006.

\bibitem{friswell2006damage}
Michael~I Friswell,
\newblock ``Damage identification using inverse methods,''
\newblock {\em Philosophical Transactions of the Royal Society A: Mathematical,
  Physical and Engineering Sciences}, vol. 365, no. 1851, pp. 393--410, 2006.

\bibitem{harley2017managing}
J.B Harley, C.~Liu, J.I Oppenheim, and J.MF Moura,
\newblock ``Managing complexity, uncertainty, and variability in guided wave
  structural health monitoring,''
\newblock {\em SICE Journal of Control, Measurement, and System Integration},
  vol. 10, no. 5, pp. 325--336, 2017.

\bibitem{raissi2019physics}
Maziar Raissi, Paris Perdikaris, and George~E Karniadakis,
\newblock ``Physics-informed neural networks: A deep learning framework for
  solving forward and inverse problems involving nonlinear partial differential
  equations,''
\newblock {\em Journal of Computational Physics}, vol. 378, pp. 686--707, 2019.

\bibitem{karpatne2017physics}
Anuj Karpatne, William Watkins, Jordan Read, and Vipin Kumar,
\newblock ``Physics-guided neural networks (pgnn): An application in lake
  temperature modeling,''
\newblock {\em arXiv preprint arXiv:1710.11431}, 2017.

\bibitem{oberkampf2004challenge}
William~L Oberkampf, Jon~C Helton, Cliff~A Joslyn, Steven~F Wojtkiewicz, and
  Scott Ferson,
\newblock ``Challenge problems: uncertainty in system response given uncertain
  parameters,''
\newblock {\em Reliability Engineering \& System Safety}, vol. 85, no. 1-3, pp.
  11--19, 2004.

\bibitem{worden2001rayleigh}
Keith Worden,
\newblock ``Rayleigh and lamb waves-basic principles,''
\newblock {\em Strain}, vol. 37, no. 4, pp. 167--172, 2001.

\bibitem{chollet2015keras}
Fran\c{c}ois Chollet et~al.,
\newblock ``Keras,'' \url{https://keras.io}, 2015.

\bibitem{srivastava2014dropout}
Nitish Srivastava, Geoffrey Hinton, Alex Krizhevsky, Ilya Sutskever, and Ruslan
  Salakhutdinov,
\newblock ``Dropout: a simple way to prevent neural networks from
  overfitting,''
\newblock {\em The Journal of Machine Learning Research}, vol. 15, no. 1, pp.
  1929--1958, 2014.

\bibitem{gal2016dropout}
Yarin Gal and Zoubin Ghahramani,
\newblock ``Dropout as a bayesian approximation: Representing model uncertainty
  in deep learning,''
\newblock in {\em in Proc. of the international conference on machine
  learning}, 2016, pp. 1050--1059.

\end{thebibliography}

\end{document}